\documentclass{article}

\usepackage{microtype}
\usepackage{graphicx}
\usepackage{subfigure}
\usepackage{booktabs} 

\usepackage{hyperref}

\usepackage[accepted]{icml2025}

\usepackage{amsmath}
\usepackage{amssymb}
\usepackage{mathtools}
\usepackage{amsthm}

\usepackage[capitalize,noabbrev]{cleveref}

\theoremstyle{plain}

\theoremstyle{definition}

\theoremstyle{remark}

\usepackage[textsize=tiny]{todonotes}

\icmltitlerunning{Refining embeddings with fill-tuning: data-efficient generalised performance improvements for materials foundation models}

\begin{document}

\twocolumn[
\icmltitle{Refining embeddings with fill-tuning: data-efficient generalised performance improvements for materials foundation models}


\begin{icmlauthorlist}
\icmlauthor{Matthew P. Wilson}{ibm}
\icmlauthor{Edward O. Pyzer-Knapp}{xyme}
\icmlauthor{Nicolas Galichet}{ibm}
\icmlauthor{Luke Dicks}{ibm,xyme}
\end{icmlauthorlist}

\icmlaffiliation{ibm}{IBM Research Europe, Hartree Centre, Daresbury, United Kingdom}
\icmlaffiliation{xyme}{Xyme, Manchester, United Kingdom}

\icmlcorrespondingauthor{Luke Dicks}{ldicks@xyme.ai}

\icmlkeywords{Machine Learning, ICML}

\vskip 0.3in
]

\printAffiliationsAndNotice{}  

\begin{abstract}
Pretrained foundation models learn embeddings that can be used for a wide range of downstream tasks. These embeddings optimise general performance, and if insufficiently accurate at a specific task the model can be fine-tuned to improve performance. For all current methodologies this operation necessarily degrades performance on all out-of-distribution tasks. In this work we present `fill-tuning', a novel methodology to generate datasets for continued pretraining of foundation models that are not suited to a particular downstream task, but instead aim to correct poor regions of the embedding. We present the application of roughness analysis to latent space topologies and illustrate how it can be used to propose data that will be most valuable to improving the embedding. We apply fill-tuning to a set of state-of-the-art materials foundation models trained on $\mathcal{O}(10^9)$ data points and show model improvement of almost 1\% in all downstream tasks with the addition of only 100 data points. This method provides a route to the general improvement of foundation models at the computational cost of fine-tuning.
\end{abstract}

\section{Introduction}

Foundational models are central to machine learning research \cite{Bommasani2021}. These models are pretrained on vast amounts of data, via self-supervised learning, enabling them to perform a variety of tasks. Many such models learn an embedding of the training data, which can be used as input for downstream tasks. The structure of this embedding, or latent space, is central to model performance, where it should encode similar points in close proximity.

Initially, foundational models were the purview of language \cite{Devlin2018, Brown2020}, but such approaches are being applied to chemistry and materials research \cite{Ahmad2022, Takeda2023}. These models have been developed for a range of modalities: text \cite{Schwaller2019, Ross2022}, three-dimensional positions \cite{Batatia2023} and molecular graphs \cite{Kishimoto2023}. Current research aims to combine these different representations into multimodal foundation models \cite{Takeda2023, Chang2024}.

Training foundation models requires both vast computational resources and datasets. For language models data is plentiful and easily accessible, but for materials applications there is significantly less data available and its generation can be both expensive and slow. Therefore, pretrained models are appealing, allowing generally good performance out of the box, and if the performance is insufficient on a given task it can be increased by fine-tuning.

Fine-tuning of pretrained models allows the development of more accurate models with a small amount of data and a greatly reduced training time \cite{Goodfellow2013}. During fine-tuning the user performs additional training with a dataset containing many targeted examples related to a given task, aiming to increase performance in this domain. The result of improved performance at the given task is degradation in all out-of-distribution tasks \cite{Luo2023}, leading to worse general performance. Therefore, many schemes have been developed that attempt to limit the degradation of fine-tuning on out-of-distribution tasks \cite{Hu2021, Mukhoti2023}. Related to fine-tuning is the field of continual learning \cite{Wang2023}, where a model continues to be updated as more data becomes available. This scenario is similar to fine-tuning, but the additional data is governed by availability and is not specific to any given task. Another related technique is continued pretraining \cite{Parmar2024}, where additional self-supervised training is performed on a pretrained model.

There is growing interest in using physical models to interpret and inform machine learning research \cite{Yang2021, Niroomand2024}. Such analyses probe properties of the loss landscapes to explain performance and develop more accurate models \cite{Draxler2018, Zhang2018, Niroomand2024_2}. However, there is little research on applying these techniques to foundation models via their latent spaces.

In this work, we present the application of physics-based roughness analysis \cite{Dicks2024} to foundation model embeddings. We demonstrate how roughness analysis can be adapted to query the quality of pretrained model embeddings, and propose a methodology to generate datasets for continued pretraining that preferentially sample from the poorest regions. We term this approach `fill-tuning', as this targeted data generation is based on model properties, not a specific task, allowing general performance improvements by filling gaps in model understanding. We illustrate this approach on a family of state-of-the-art materials foundation models \cite{Priyadarsini2024} trained on up to 8B data points, where the injection of only 100 data points into regions where the embedding is poorest allows continued training to give almost 1\% improvement across all downstream tasks.

\section{Methods}

\subsection{Foundation Models}

Roughness analysis has been widely used to relate the topology of datasets to the performance of machine learning models \cite{Aldeghi2022, Graff2023}. Here, we extend the frustration metric \cite{Dicks2024}, previously showing state-of-the-art performance in predicting model performance from dataset structure, to foundation model latent spaces.

We expect that, after successful training, a foundation model embedding will have placed similar molecules in close proximity, and such a property is essential for methods that leverage latent space interpolation \cite{Bombarelli2018}. This condition is similar to that studied in quantitative structure-property relationships where regions of feature space that exhibit large changes in molecular similarity over small distances, termed activity cliffs \cite{Stumpfe2019}, degrade model performance. Therefore, we assume that the smoothness of properties (herein, the molecular similarity within the embedding) can be a proxy for the accuracy of downstream tasks trained from the embedding.

We apply roughness analysis to the latent space of the SELFIES-TED\footnote{The model SELF-BART is now referred to as an earlier iteration of the SELFIES-TED model.} materials foundation models \cite{Priyadarsini2024}, a family of models based on the BART architecture \cite{Lewis2019} and trained on molecular SELFIES \cite{Krenn2020} strings. There are three variants of this model, which we distinguish by the number of parameters and training data size. In all cases data is drawn from both ZINC22 \cite{Tingle2023} and PubChem \cite{Kim2016}. The models, and their names, are summarised in Table~\ref{foundation-models}.

\begin{table}[t]
\caption{Summary of the various SELFIES-TED materials foundation models studied in this work.}
\label{foundation-models}
\vskip 0.15in
\begin{center}
\begin{small}
\begin{sc}
\begin{tabular}{lccc}
\toprule
Model & Parameters & Dataset size \\
\midrule
SELFIES-TED (small) & 2.2M & 8B \\
SELFIES-TED (medium)  & 358M & 500M \\
SELFIES-TED (large) & 358M & 1B \\
\bottomrule
\end{tabular}
\end{sc}
\end{small}
\end{center}
\vskip -0.1in
\end{table}

We perform our initial analysis on the SELFIES-TED (small) model, and subsequent analysis on the larger models. The SELFIES-TED (small) model consists of a transformer encoder-decoder pair. The encoder produces a 256-dimensional embedding per token, which we project onto a 128-dimensional embedding for the whole sequence. We compute the roughness over this 128-dimensional embedding.

\subsection{Frustration Analysis}

To apply the frustration metric we require a continuous surface that can be queried at all points in the embedding. We construct such a surface by associating a neighbourhood molecular similarity to each target point in the embedding. Starting from each target point, we add 10 uniformly sampled vectors with an $L_2$ norm of 0.05 and decode each resulting point in the embedding to SELFIES. From the SELFIES we compute the mean Tanimoto similarity between each neighbour and the target point, after conversion to Morgan fingerprints \cite{Morgan1965}. For computational efficiency we compute the molecular similarity at 5000 points in the embedding, chosen via Latin-hypercube sampling and infer the complete property surface via radial basis function interpolation \cite{Hardy1971} with a thin-plate kernel. We specify the smoothness as $10^{-5}$ throughout to ensure a faithful description of the dataset.

Given the continuous similarity surface constructed for the latent space we perform the frustration analysis. We decompose the surface into its stationary points, which are distinguished as either minima (only positive eigenvalues of the second derivative matrix) or transition states (a single negative eigenvalue of the second derivative matrix). All minima were located using random initialisation and minimisation. Transition states were located between the set of local minima using the nudged elastic band \cite{Henkelman2000} and hybrid-eigenvector following \cite{Munro1999} algorithms. In contrast to previous work, we directly explore the 128-dimensional space to locate its stationary points.

\begin{figure}[ht]
\vskip 0.2in
\begin{center}
\centerline{\includegraphics[width=\columnwidth]{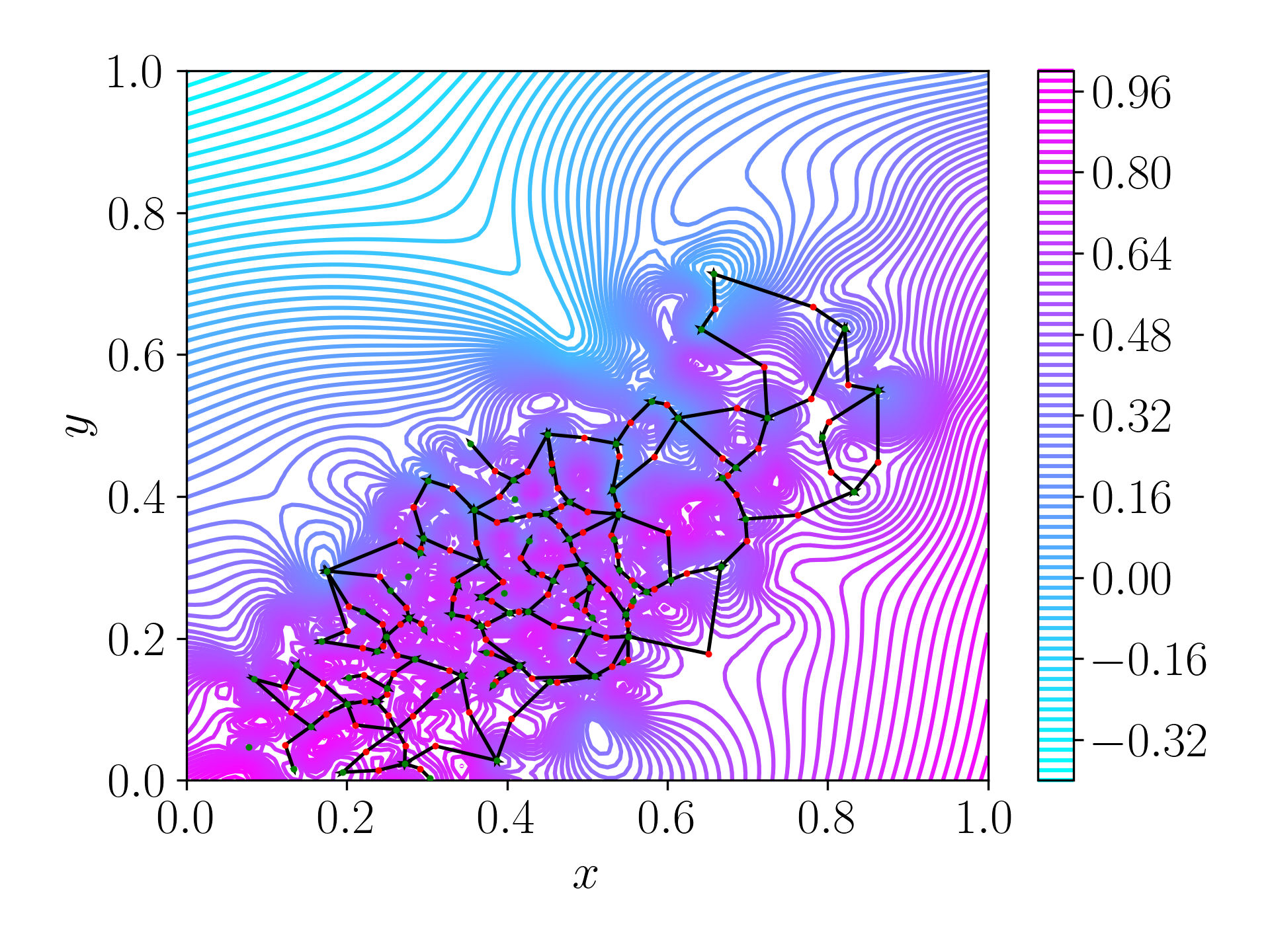}}
\centerline{\includegraphics[width=0.87\columnwidth]{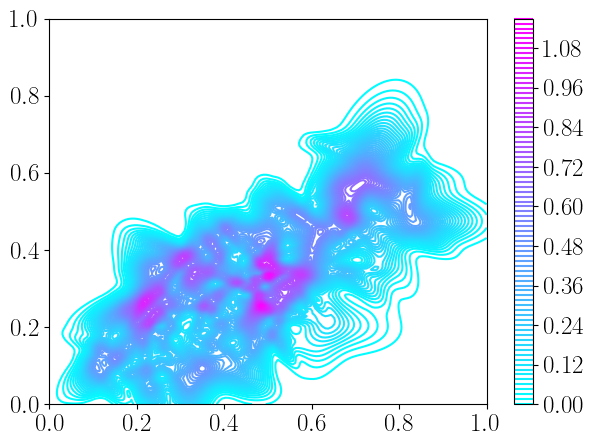}}
\caption{An illustrative kinetic transition network constructed from a two-dimensional molecular similarity measure (top). Local minimum (green) and transition states (red) are connected by solid lines when they are joined by steepest-descent paths. The corresponding continuous roughness surface is constructed from a sum of multivariate normals (bottom).}
\label{ktn}
\end{center}
\vskip -0.2in
\end{figure}

Each transition state is connected to the two minima obtained by steepest-descent paths along the eigenvector corresponding to the negative eigenvalue. Therefore, the set of transition states and their connected minima can be conveniently represented as a weighted graph, known as a kinetic transition network \cite{Noe2008}. In such a graph, each minimum is a node and edges exist between any pair of minima directly connected by a transition state. An example kinetic transition network is given in Fig.~\ref{ktn}.

Given a kinetic transition network constructed for the similarity surface we can associate roughness with edges of the graph. The frustration associated with each particular direction from minimum to transition state, $F_{i}$, is given by
\begin{equation}
    F_{i} = \mathrm{exp} \left( - \frac{d(\mathbf{x}_{\mathrm{min}}, \mathbf{x}_\mathrm{ts})^2}{2l^2} \right) \cdot \left(f_{\mathrm{ts}} - f_{\mathrm{min}}\right).
\end{equation}
The first term provides an appropriate reweighting of the frustration contributions based on the proximity of minimum and transition state at $\mathbf{x}_{\mathrm{min}}$ and $\mathbf{x}_\mathrm{ts}$, respectively. $d(\mathbf{x}_i, \mathbf{x}_j)$ is the Euclidean distance and $l$ is the lengthscale; specified as 80.0 in this work. The second term denotes the change in function value between the two stationary points. Large contributions from either of these terms indicates high roughness.

\subsection{Roughness Sampling} \label{roughness-sampling}

In previous work, frustration was averaged over the whole kinetic transition network \cite{Dicks2024}. This overall frustration metric correlates strongly with the error obtained after training a machine learning model on the dataset. However, our aim here was instead to construct a dataset that samples from regions of high roughness.

Regions of high roughness occur where the molecular similarity is most rapidly varying within the embedding. Such changes will occur at regions of the embedding that have high molecular similarity, proximal to those that have low similarity. Roughness analysis will specifically highlight where the quality of the embedding is most rapidly decreasing, which will likely lie within, or at the edges of, the space spanned by the encoded training data; the most relevant regions for downstream applications. Injection of data into these rough regions may allow further training to refine the embedding and mitigate performance degradation by providing relevant examples where they can exert the most influence on model performance.

The frustration is associated with vectors in the latent space and we transform these into a continuous roughness surface via a sum of weighted Gaussians. We place a multivariate Gaussian along each vector from minimum to transition state, and sum them according to
\begin{equation}
    \mathcal{F} = \sum_{i} F_i \mathcal{N}(\boldsymbol{\mu} = \mathbf{x}^{\mathrm{min}}_i + \frac{3}{4}(\mathbf{x}^{\mathrm{ts}}_i - \mathbf{x}^{\mathrm{min}}_i) | \boldsymbol{\Sigma} = \mathbf{R}_i \boldsymbol{\Sigma}_i \mathbf{R}_i^T)
\end{equation}
where
\begin{equation}
    \boldsymbol\Sigma_i = \begin{pmatrix}
               \sigma_i & 0 & 0 \\
               0 & \delta_{ij} & 0 \\
               0 & 0 & \delta_{ik}
               \end{pmatrix}.
\end{equation}
$\mathbf{R}$ is the rotation matrix that transforms the basis such that the first dimension of the Gaussian lies along the corresponding frustration vector. $\sigma$ controls the decay of the distribution along this direction between minimum and transition state, which is fixed at 0.99 throughout. $\delta$ is a constant value to control the variance in all orthogonal directions, set to 0.25 in this work. The scaling of each multivariate normal by $F_i$ ensures that their contribution is proportional to the corresponding local frustration. From this surface we locate the 100 highest minima by basin-hopping \cite{Wales1997}, which we decode to form our fill-tuning dataset.

We continue training of the base model from its final checkpoint; as in the original training protocol we randomly mask 15\% of the input tokens and the labels are the original SELFIES. For continued training we provide only the small fill-tuned dataset (100 data points) and train for up to 50 epochs, taking the best-performing checkpoint when considering both training and validation loss.

We term this process fill-tuning to distinguish from the usual fine-tuning workflow, wherein the dataset used to continue training is designed to cater to a particular task, and consequently leads to degraded performance on other tasks. Instead, datasets generated by fill-tuning are not designed to cater to a particular task, but to influence regions where we estimate the embedding is poorest. Consequently, our aim is to provide the most relevant data for improving the overall state of the pretrained model, allowing it to perform better in all downstream tasks.

\section{Results}

\subsection{Direct Fill-Tuning} \label{direct-fill-tuning}

We illustrate the effect of fill-tuning on the family of SELFIES-TED materials foundation models. We determine the performance of both base and fill-tuned models through construction of classification models from the embedding using XGBoost \cite{Chen2016}, as in \cite{Priyadarsini2024}. We train models for classification of the BACE, BBBP, ClinTox and HIV tasks, as contained in the MoleculeNet benchmark tests \cite{Wu2018}. The performance of each model is given by the ROC-AUC score. The score across all the benchmarks, which probe different parts of chemical space, is indicative of embedding quality. Good performance in all tests results from a high quality embedding, in which similar molecules have been encoded in close proximity.

We present results for each classification model trained from the embeddings of the base model, SELFIES-TED (small), and the corresponding model generated by fill-tuning, in Table~\ref{classification-tasks}. For reference, we evaluate several alternative continued pretraining approaches applied to the same base model.
\begin{itemize}
    \item \textit{Random} -- We trained the model using 100 random points in the embedding, decoded to SELFIES. This benchmark reflects the selection of additional data not specific to a given task, but not selected based on latent space properties.
    \item \textit{BACE} -- We continued training of the base model using 100 samples taken from the training data for one classification task (BACE). This dataset mimics a standard fine-tuning dataset. 
    \item \textit{Combined} -- We continued training of the base model using 100 samples from the training data of each of the four classification tasks, combined into a single dataset. 
    \item \textit{Sequential} -- We continued training of the base model with the same 100 samples for each task as above, but sequentially in four separate training runs.
\end{itemize}
The size of these baseline datasets are designed to match that of fill-tuning, but without any consideration of the latent space structure.

\begin{table*}[t]
\caption{Classification accuracies for benchmark tasks taken from MoleculeNet. We evaluate the performance of XGBoost models trained from the embedding of the base pretrained model, and those obtained after several continued pretraining baselines and fill-tuning. The performance is given by the ROC-AUC, and the percentage change from the base model is given in parentheses.}
\label{classification-tasks}
\vskip 0.15in
\begin{center}
\begin{small}
\begin{sc}
\begin{tabular}{lccccccr}
\toprule
Task & Base model & Random & BACE & Sequential & Combined & Fill-tuned \\
\midrule
BACE    & 0.8540 & 0.8558 (+0.2) & 0.8573 (+0.4) & 0.8705 (+1.9) & 0.8448 (-1.1) & \bf{0.8742} (+2.4) \\
BBBP    & 0.9057 & 0.9073 (+0.2) & 0.9000 (-0.6) & 0.8953 (-1.2) & 0.9064 (+0.1) & \bf{0.9112} (+0.6) \\
ClinTox & \bf{0.8805} & 0.8751 (-0.6) & 0.8391 (-4.7) & 0.8544 (-3.0) & 0.7994 (-9.2) & 0.8763 (-0.5) \\
HIV     & 0.7589 & 0.7553 (-0.5) & 0.7569 (-0.3) & \bf{0.7677} (+1.2) & 0.7623 (+0.5) & 0.7627 (+0.5) \\
\bottomrule
\end{tabular}
\end{sc}
\end{small}
\end{center}
\vskip -0.1in
\end{table*}

It is worth noting that the corresponding foundation model is trained on $\mathcal{O}(10^9)$ data points. Therefore, the addition of 100 data points constitutes a tiny fraction (0.00001\%) of the data the model has observed when constructing the initial embedding. Continued pretraining with randomly-selected embeddings results in a small reduction in performance across all downstream tasks. The emphasis placed on random data, likely far from the training data, degrades performance in the region of the embedding relevant for the classification tasks. Moreover, adding data specific to the evaluation tasks, by continued pretraining with training examples from the MoleculeNet tasks, also degrades general performance. All forms of such continued pretraining can lead to improved performance on their specific task, but the performance degradation in all other tasks limits applicability for multi-task optimisation. Even with sequential training, and training data from all tasks, the performance decrease across multiple tasks can be severe.

However, fill-tuning shows an average increase of 0.75\% across all downstream tasks. The base pretrained model already produces downstream models of high accuracy, and with fill-tuning we can extract additional general performance with only 100 targeted data points. These results illustrate that fill-tuning does not bias to one particular task at the expense of others. Instead, we see a meaningful average improvement in the classification tasks, indicating that additional training is able to improve relevant regions of the pretrained model embedding. Due to its size, such a dataset will be fast to generate, even with expensive simulations or experiments, allowing improvement of foundation model performance at low cost.

\begin{figure}[h!]
\vskip 0.2in
\begin{center}
\centerline{\includegraphics[width=0.9\columnwidth]{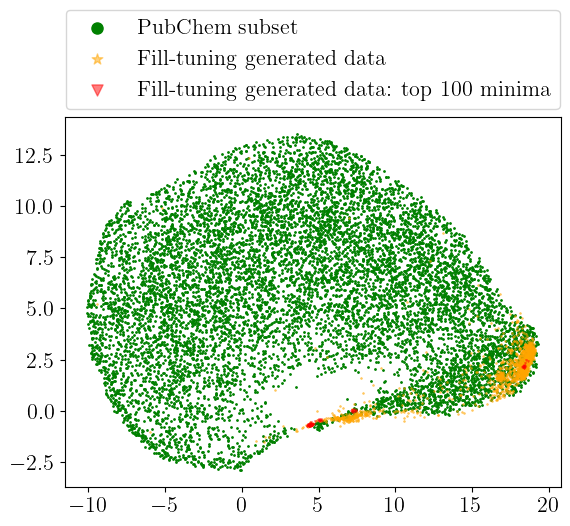}}
\centerline{\includegraphics[width=0.9\columnwidth]{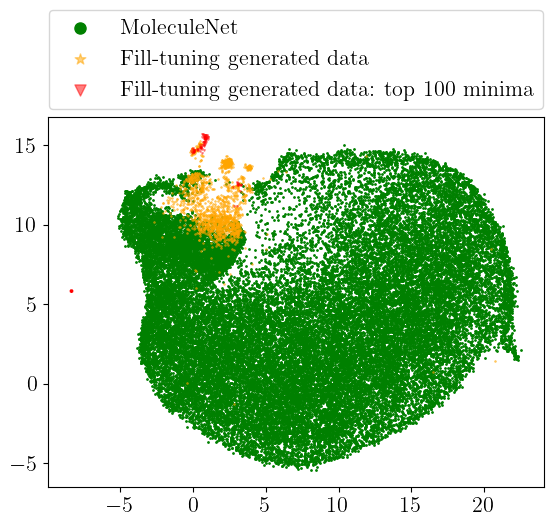}}
\caption{UMAP projections of the SELFIES-TED (small) latent space. We fit a UMAP reducer using embeddings of $10^4$ molecules sampled from PubChem (top) and all the training data from MoleculeNet classification tasks (bottom), both of which are combined with the data generated by fill-tuning.}
\label{umap}
\end{center}
\vskip -0.2in
\end{figure}

An analysis of the fill-tuning dataset is insightful and we present visualizations of the SELFIES-TED (small) latent space in Figure \ref{umap}. We used UMAP \cite{Mcinnes2020} to generate a two-dimensional projection of the embeddings of $10^4$ molecules randomly sampled from PubChem, which approximately captures the regions of the foundation model embedding that encode its training data. We jointly projected the embeddings generated by fill-tuning. It is striking how few of the molecules generated by fill-tuning exhibit standard chemistry and we observe that they overlap largely at the edges of the two-dimensional distribution, in agreement with the expectations of Sec.~\ref{roughness-sampling}. Many of these points selected by fill-tuning, for this projection, lie within gaps in the training data, which may evidence that the method is filling knowledge gaps even within regions spanned by the training data. Moreover, a significant portion of the fill-tuning data lies slightly outside the space enclosed by training data, where the model performance is beginning to degrade due to a lack of information. Data in this region could help to extend where the model can be accurately applied.

We perform the same analysis on the training data for the MoleculeNet classification tasks. Here, the projection relates the fill-tuning data to the data used for evaluating model performance. We observe that the fill-tuning data is highly distinct from that used in the classification benchmarks. The addition of fill-tuning data, although in distinct regions of chemical space to that required for classification tasks, must allow favourable reconfiguring of the embedding within these regions. Such learning is akin to language development in children, where reading of nonsense words can help generalise phonics skills to new (real) words. A related phenomenon has been observed in which invalid SMILES strings remain valuable for developing accurate models for regions of chemical space corresponding to valid SMILES strings \cite{Skinnider2024}.

\subsection{Indirect Fill-Tuning}

The set of SELFIES-TED models all have different, but related, architectures, along with a varying number of parameters and training data. However, all models use training data from the same set of sources. Therefore, we evaluate whether fill-tuning can be performed on one model and indirectly applied to related models. The models become increasingly performant, as measured by the benchmark tests, from small to large. Therefore, if successful, fill-tuning can be applied to smaller models, at reduced computational cost, and leveraged to increase performance of state-of-the-art models.

We use the same fill-tuning dataset as in the previous section, derived from the latent space of the SELFIES-TED (small) model, and perform continued training of the larger models with it. We illustrate the performance of each original pretrained model, and its fill-tuned equivalent, for the benchmark classification tasks in Table \ref{larger-models}. Surprisingly, we continue to see performance gains across the set of validation tasks for all larger models.

\begin{table*}[t]
\caption{Accuracies of XGBoost models trained from different model embeddings. We compare the performance of the base model with that after 50 epochs of fill-tuning for a range of molecular classification tasks. We use the same dataset of 100 data points in all additional training. We report the ROC-AUC score for each task, and the percentage change is given in brackets for the fill-tuned models relative to the original.}
\label{larger-models}
\vskip 0.15in
\begin{center}
\begin{small}
\begin{sc}
\begin{tabular}{lcccccc}
\toprule
 & \multicolumn{2}{c}{SELFIES-TED (small)} & \multicolumn{2}{c}{SELFIES-TED (medium)} & \multicolumn{2}{c}{SELFIES-TED (large)} \\
Task & Base & FT & Base & FT & Base & FT \\
\midrule
BACE    & 0.8540 & 0.8742 (+2.4) & 0.8628 & 0.8483 (-1.7)  & 0.8505 & 0.8793 (+2.9) \\
BBBP    & 0.9057 & 0.9112 (+0.6) & 0.9095 & 0.9207 (+1.2) & 0.9117 & 0.9193 (+0.8) \\
ClinTox & 0.8805 & 0.8763 (-0.5) & 0.8586 & 0.8680 (+1.1) & 0.9243 & 0.9213 (-0.3) \\
HIV     & 0.7589 & 0.7627 (+0.5) & 0.7892 & 0.7923 (+0.4) & 0.8086 & 0.8132 (+0.5) \\ \hline
Av. Percentage change & & +0.75 & & +0.25 & & +0.98 \\ 
\bottomrule
\end{tabular}
\end{sc}
\end{small}
\end{center}
\vskip -0.1in
\end{table*}

In the largest, and most performant model, we see an even more significant increase in performance. This model already has state-of-the-art performance and with fill-tuning we extract an additional 1\% performance across all downstream tasks.

The ability for the same fill-tuning dataset to improve performance of all models highlights that similar deficiencies arise in all models. The addition of more parameters, and updated architectures, does not sufficiently change the structure of the embedding in its poorest regions. Therefore, these results highlight that increased data quantity may not improve model quality. Instead, data character is more important to producing foundation models with better performance, and data of the appropriate character can be determined by roughness analysis.

\section{Conclusions and Future Work}

Here we summarise the key developments and findings of this work:
\begin{itemize}
    \item We present fill-tuning, a novel method for generating datasets for continued pretraining that are not specific to a task, but to gaps in model understanding.
\end{itemize}
In general, fine-tuning is designed to increase performance at a specified task. Standard fine-tuning datasets contain repeated examples relevant to the given task, which results in reduced performance on out-of-distribution tasks. With fill-tuning, we generate datasets without reference to downstream tasks, as our aim is not to increase performance in any specific task, but to improve regions of the embedding. Consequently, further training can make a foundation model generally more performant at all downstream tasks.

\begin{itemize}
    \item We illustrate that with the application of fill-tuning it is possible to improve the general performance of foundation models with a tiny amount of data.
\end{itemize}
In many applications we consider the pretrained model to be the optimum of general performance, and fine-tuning is used to increase performance in only a specific task. Here, we illustrate that even after pretraining with up to 8B data points it is possible to improve general foundation model performance with selective data generation. The amount of data required can be vanishingly small compared to the training data, allowing a short timescale for foundation model improvement. This result provides evidence that data quantity alone may be an insufficient predictor of foundation model performance.

\begin{itemize}
    \item The fill-tuning data that improves model performance has little relevance to the benchmark tasks, and is distinct from the foundation model training data.
\end{itemize}

The data generated by fill-tuning contains examples that are outside the foundation model training data, despite its size. The fill-tuning dataset also belongs to vastly different regions of chemical space to those evaluated in the classification tasks and, yet, it still permits model improvement. This result provides strong evidence that the added data reconfigures regions of the embedding responsible for poor performance, and the most appropriate data for that task does not have to correspond to standard regions of chemical space.

\begin{itemize}
    \item We illustrate that the application of fill-tuning to smaller models can permit improvement of related models with more parameters.
\end{itemize}

The data that was most necessary to correct problems in the latent space of the small model remain important in improving the embedding of the larger models. This result highlights that the additional parameters and changing architecture, although improving model performance, do not correct some persistent problems in the embedding. Therefore, the models, all constructed from the same datasets, can all be improved by the same fill-tuning dataset. Generally, such an analysis paves the way to improving state-of-the-art foundation models with less computational effort, by analysing their smaller variants.

This work provides an initial exploration of foundation model latent space topology and its application to continued pretraining. The given example provides clear evidence that latent space topology can be leveraged to build more performant materials foundation models. There are many avenues for further research, and extensions to alternative foundation model architectures and modalities. Therefore, we hope this methodology constitutes a step on the pathway to more advanced fine-tuning methods and, ultimately, more accurate foundation models.

\section*{Software and Data}

The topsearch Python package \cite{topsearch} was used to generate all results in this work, which is freely available at \url{https://github.com/IBM/topography-searcher}. Example scripts to reproduce the work herein are included within the Github repository at \url{https://github.com/IBM/topography-searcher/tree/main/examples/scripts/dataset_roughness/latent_space_roughness}.

\section*{Acknowledgements}

The authors would like to acknowledge the financial support of the Hartree National Centre for Digital Innovation – a collaboration between the Science and Technology Facilities Council and IBM.

\bibliography{main}
\bibliographystyle{icml2025}

\end{document}